%% file: main.tex
\documentclass[10pt,twocolumn,letterpaper]{article}

\usepackage{iccv}              

\usepackage{multirow}
\usepackage{comment}
\usepackage{colortbl}

\definecolor{iccvblue}{rgb}{0.21,0.49,0.74}
\usepackage[pagebackref,breaklinks,colorlinks,allcolors=iccvblue]{hyperref}

\def\paperID{***}
\def\confName{***}
\def\confYear{2025}

\title{Follow Your Motion: A Generic Temporal Consistency Portrait Editing Framework with Trajectory Guidance}

\author{Haijie Yang, Zhenyu Zhang, Hao Tang, Jianjun Qian, Jian Yang}

\begin{document}
\twocolumn[{
\renewcommand\twocolumn[1][]{#1}
\maketitle
\begin{center}
    \captionsetup{type=figure}
    \includegraphics[width=0.95\textwidth]{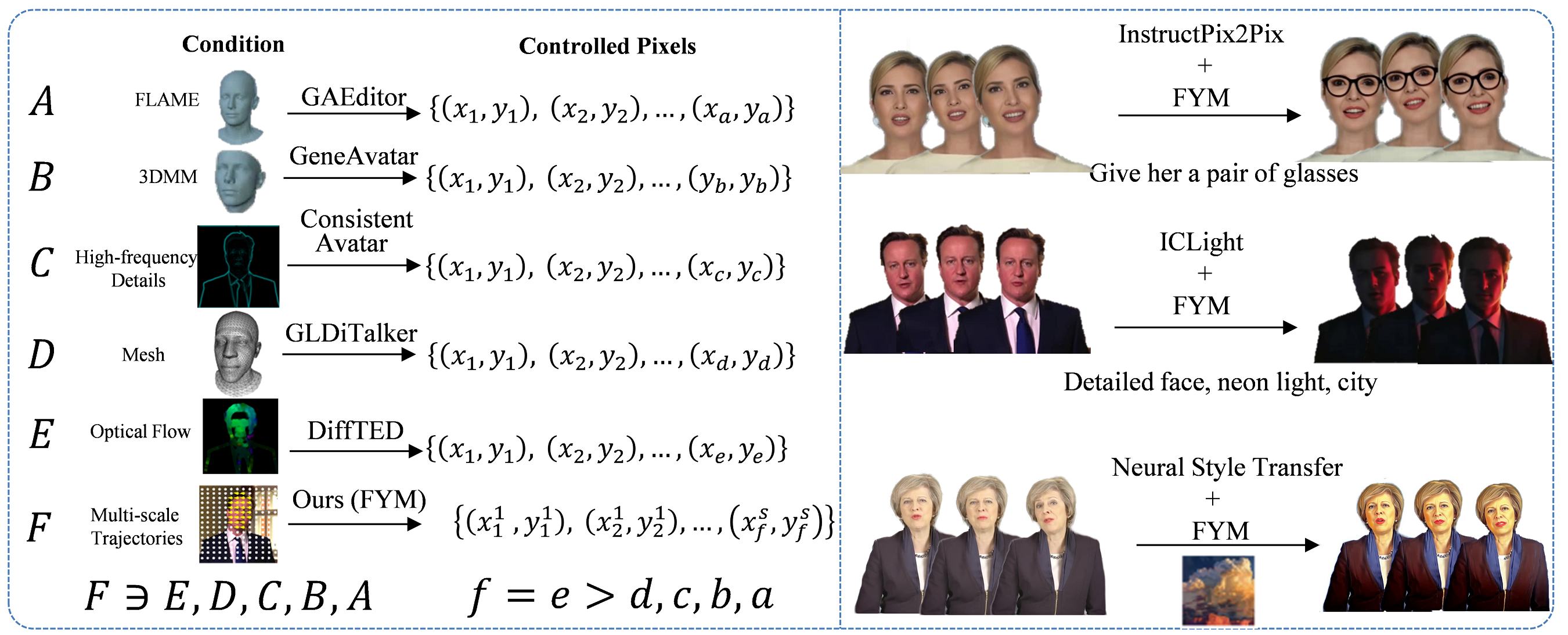}
    \captionof{figure}{Compared to other state-of-the-art 2D portrait generation methods, our approach directly controls the motion of the edited result by using multi-scale motion trajectories of pixel points as guidance, ensuring temporal consistency. The other methods in the figure can be seen as special cases of our approach. Additionally, our method supports various editing tools and optimizes the temporal consistency of the results produced by these tools.}
    \label{fig:teaser}
\end{center}
}]

\input{0_abstract}    
\input{1_intro}

\input{2_related}
\input{3_pre}

\input{4_method}
\input{5_experiment}
\input{6_discussion}
\input{7_conclusion}
{
    \small
    \bibliographystyle{ieeenat_fullname}
    \bibliography{main}
}


\end{document}

%% file: 0_abstract.tex
\begin{abstract}
Pre-trained conditional diffusion models have demonstrated remarkable potential in image editing. However, they often face challenges with temporal consistency, particularly in the talking head domain, where continuous changes in facial expressions intensify the level of difficulty. These issues stem from the independent editing of individual images and the inherent loss of temporal continuity during the editing process. In this paper, we introduce Follow Your Motion (FYM), a generic framework for maintaining temporal consistency in portrait editing.
Specifically, given portrait images rendered by a pre-trained 3D Gaussian Splatting model, we first develop a diffusion model that intuitively and inherently learns motion trajectory changes at different scales and pixel coordinates, from the first frame to each subsequent frame. This approach ensures that temporally inconsistent edited avatars inherit the motion information from the rendered avatars. Secondly, to maintain fine-grained expression temporal consistency in talking head editing, we propose a dynamic re-weighted attention mechanism. This mechanism assigns higher weight coefficients to landmark points in space and dynamically updates these weights based on landmark loss, achieving more consistent and refined facial expressions. Extensive experiments demonstrate that our method outperforms existing approaches in terms of temporal consistency and can be used to optimize and compensate for temporally inconsistent outputs in a range of applications, such as text-driven editing, relighting, and various other applications.
\end{abstract}

%% file: 1_intro.tex
\section{Introduction}
\label{sec:intro}
In virtual reality and related applications, the demand for animatable 3D talking heads has grown significantly. Methods \cite{imavatar,insta,diffusionrig,splattingavatar} like SplattingAvatar \cite{splattingavatar} have exhibited remarkable success in this area. Therefore, editing the rendered results of these talking heads is significantly valuable.

Early works \cite{lin20223dganinversioncontrollable, cai2022pix2nerfunsupervisedconditionalpigan, sun2022ide3dinteractivedisentangledediting} mainly use Generative Adversarial Networks (GANs) \cite{gan} for editing or stylizing animations based on style labels or reference images. For example, IDE-3D \cite{sun2022ide3dinteractivedisentangledediting} introduces a hybrid GAN inversion method and a canonical editor, which initializes latent codes with semantic and texture encoders and generates high-quality editing results by manipulating semantic masks. However, these methods are limited to editing preset attributes. Recently, diffusion models \cite{ddpm} have demonstrated stronger generative abilities compared to GANs. Based on denoising diffusion processes, many generative models, adapters, and fine-tuning methods \cite{instructpix2pixl,anyv2v,tokenflow} have been proposed, showing higher quality and flexibility in image generation. For example, InstructPix2Pix \cite{instructpix2pixl} combines GPT-3 \cite{GPT3} and Stable Diffusion \cite{sd} to generate training data, enabling the model to edit images according to specified instructions. However, when editing portrait videos, these methods still face challenges in maintaining temporal consistency across frames and struggle with ensuring smooth transitions between consecutive frames.

To address these issues, numerous methods \cite{consistentavatar,portrait,geneavatar,gaussianavatareditor,diffted,glditalker} with diverse forms have emerged in recent years. DiffTED \cite{diffted} leverages the learned 2D keypoints of the Thin-Plate Spline (TPS) motion model \cite{thinplate} and optical flow to ensure temporal consistency and diverse gestures. However, this sparse full-body keypoint system cannot precisely control facial motion. PortraitGen \cite{portrait} and GaussianAvatarEditor \cite{gaussianavatareditor} use the FLAME \cite{flame} facial prior model to create a dynamic 3D Gaussian field, ensuring both structural and temporal consistency across frames. Additionally, PortraitGen performs fine-grained optimization using facial landmarks. In addition, GeneAvatar \cite{geneavatar} designs a novel expression-aware modification generative model that enables lifting 2D edits from a single image to a consistent 3D modification field, which can be universally applied to various 3DMM-driven volumetric head avatars. GLDiTalker \cite{glditalker} addresses the modality misalignment issue by leveraging facial meshes and a graph latent diffusion transformer, enabling the generation of temporally stable 3D facial animations. ConsistentAvatar \cite{consistentavatar} proposes using diffusion models to learn high-frequency information of the face, including mask contour details and fine wrinkles, to maintain consistency in the generated results. However, these face-prior-based methods still face challenges in detail expression and individual adaptability. 

Based on the discussion above, we pose the following question: \textbf{Is it possible to construct a generic and effective pipeline at the most fundamental level to address the issue of temporal inconsistency in talking head editing?} We observe that \textbf{FLAME}, \textbf{3DMM}, \textbf{optical flow}, \textbf{high-frequency detail}, \textbf{mask}, \textbf{mesh}, and \textbf{landmark} represent different forms for manipulating specific pixels or regions in an image or video. For instance, FLAME and 3DMM alter facial shape by adjusting facial meshes or key points, which in turn affects the pixels in the corresponding areas of the image. Similarly, optical flow estimates the displacement of pixels over time, allowing for dynamic adjustments (see Fig. \ref{fig:teaser} for a clearer explanation). Therefore, maintaining temporal consistency in the edited portrait can be viewed as the problem of preserving the same motion trajectory as the original video—specifically, how to precisely control the pixel motion trajectory of the edited portrait.

To this end, we propose "Follow Your Motion " (FYM), a generic framework for maintaining temporal consistency in portrait editing. Specifically, our framework consists of two stages. Stage 1: We first employ 3D Gaussian Splatting (3DGS) for consistent and efficient rendering. Stage 2: We develop a diffusion model that intuitively and inherently learns the motion trajectory changes at different scales and pixel coordinates from the original video frames. We use multi-resolution hash encoding to encode pixel coordinates and compute the difference between the encoding results of each frame and the first frame. Furthermore, to maintain fine-grained expression temporal consistency in portrait editing, we propose a dynamic re-weighted attention mechanism. This mechanism assigns higher weight coefficients to landmark coordinates in space and dynamically updates these weights based on landmark loss, achieving more consistent and refined facial expressions.

In summary, our contributions are as follows:

\begin{itemize}
\item We propose FYM, a generic framework for maintaining temporal consistency in portrait editing.

\item We design a diffusion model that learns the motion trajectory changes from the original video frames, thereby constraining the motion of the edited portrait to achieve overall content temporal consistency.

\item We present a dynamic re-weighted attention mechanism that assigns higher weight coefficients to landmark coordinates in space, ensuring more refined facial expressions and fine-grained temporal consistency.

\end{itemize}
\begin{figure*}[ht!]
    \centering
    \includegraphics[width=\textwidth]{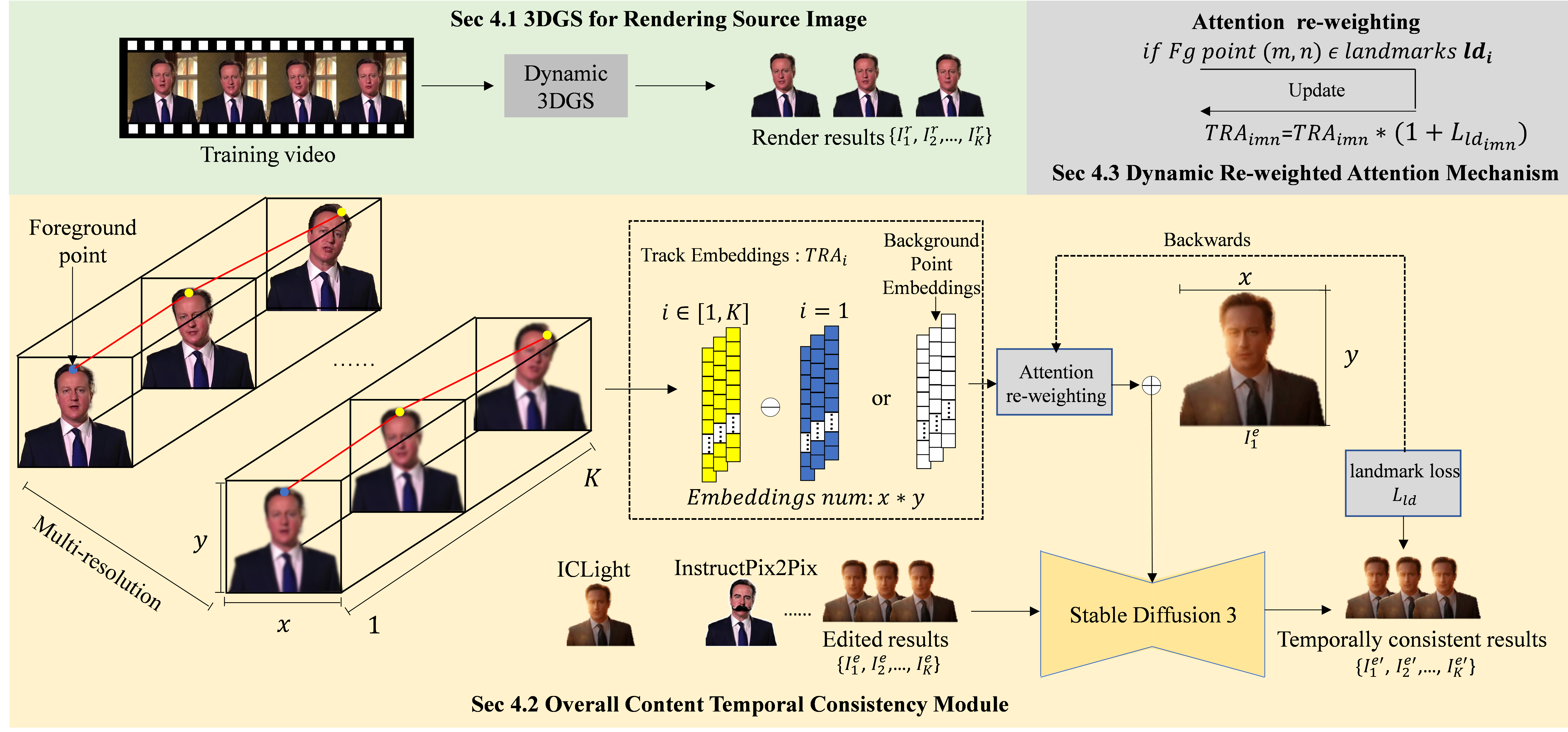}
    \caption{
    \textbf{Overview}. FYM begins with an efficient 3DGS model to render temporally consistent portraits (Sec \ref{sec:3dgs}). Next, we develop a diffusion model that intuitively and inherently learns the motion trajectories changes at different scales and pixel coordinates from the original video frames (Sec \ref{sec:OCTCM}). Finally, we propose a dynamic re-weighted attention mechanism. This mechanism assigns higher weight coefficients to landmark coordinates in space and dynamically updates these weights based on landmark loss, achieving more consistent and refined facial expressions (Sec \ref{sec:dram}).}
    \label{fig:pipeline}
\end{figure*}

%% file: 2_related.tex
\section{Related Work}
\label{sec:related}
\textbf{Talking Head Representation.} Early methods \cite{sadtalker,ren2021pirenderer,discofacegan} primarily use 3DMM \cite{3dmm} priors to guide the generator. For example, PIRenderer \cite{ren2021pirenderer} integrates 3DMM with speech information as action descriptors to generate new actions from a single image. Recently, researchers have combined 3D representation techniques with 3DMM to achieve more precise control over the expressions and poses of characters. NHA \cite{nha} and IMAvatar \cite{imavatar} enhance the mesh topology of FLAME \cite{flame}, producing more realistic mesh-based avatars. Additionally, INSTA \cite{insta} and RigNeRF \cite{rignerf} use 3DMM to build radiance fields, enabling more detailed facial representations. Although these methods have produced promising rendering results, their rendering efficiency remains limited. Recently, 3D Gaussian Splatting (3DGS) \cite{3dgs} has been introduced to digital head modeling. With its flexible representation and efficient, differentiable rasterizer, 3DGS has made significant strides in modeling efficiency and rendering fidelity, highlighting its potential for practical applications. In this work, we employ the highly efficient GaussianSplatting \cite{splattingavatar} in the first stage to obtain high-quality, editable renderings.

\noindent\textbf{Diffusion Model for Image/Video Generation and Editing.} Diffusion models \cite{ddpm,ddim} have become a key component in the text-to-image (T2I) field. For instance, Imagen \cite{imagegen} generate high-resolution images using cascaded diffusion models. Later, Stable Diffusion \cite{sd} introduced the idea of training diffusion models in a learned latent space, reducing computational complexity. Early image editing techniques \cite{spatext,makeasense} required users to provide additional masks, which was time-consuming. To address this limitation, methods such as InstructPix2Pix \cite{instructpix2pixl} allow for editing based solely on textual instructions from users. To further enhance controllability, ControlNet \cite{controlnet} adds an extra branch to Stable Diffusion, enabling the input of various control signals, such as depth and normal. In the field of text-to-video (T2V), several notable works \cite{scg,modelscope,align} leverage Stable Diffusion to generate high-quality videos through learned latent spaces. For text-driven video editing, Tune-A-Video \cite{tuneavideo}  introduces an efficient tuning strategy based on Stable Diffusion, while subsequent methods \cite{Pix2Video,FLATTEN,256base,tokenflow}, develop various attention mechanisms for zero-shot editing to capture temporal cues. Additionally, AnyV2V \cite{anyv2v} adopts an image-driven approach for editing, but fast motion in the image can lead to blurry video outputs. Furthermore, text-driven video editing methods often struggle with fine-grained local edits, such as modifying hair. In our work, conventional video editing methods are inadequate for talking-head videos, as they typically involve rapid changes in pose and expression. Portraits, too, demand fine-grained editing. To address these challenges, we propose a novel and intuitive framework specifically designed for talking-head videos.

\noindent\textbf{Portrit Editing.} The editing of portrait appearance and semantic attributes plays a crucial role in the application of virtual communities. Early efforts \cite{firstorder, sparse, stylerig, analyzing, flnet,sadtalker} focused on using pre-trained GAN models for facial editing. However, these models have inherent limitations, as they rely on a single-image 2D generation process. This approach results in complex image rendering and limited facial fitting capabilities. To overcome these challenges, some methods \cite{clipface,tig} employ 3D representations, such as 3DMM, as geometric proxies to improve 3D consistency during editing. However, the use of mesh models still restricts the ability to produce personalized results. Other methods, such as NeRF, have been used for editing, but they are hindered by inefficiencies. Recently, diffusion models have gained traction for portrait editing. TextDeformer \cite{textdeformer} and Texture \cite{texture} modify geometry and texture based on text prompts, respectively. PhotoMaker \cite{photomaker} combines multiple ID images to construct stacked inputs and uses text prompts for image-to-image style transfer. In contrast, many studies leverage diffusion models and ControlNet to explore controllable face generation. In our work, we also employ ControlNet to achieve consistent editing results.

%% file: 3_pre.tex
\section{Preliminary}
\label{sec:pre}
\textbf{3D Gaussian Splatting.} Gaussian Splatting uses point clouds to model a scene using a set of 3D Gaussians. Each Gaussian ellipse is defined by its color $c$, opacity $o$, and position $x$, with the covariance matrix $\Sigma$ described by the equation: $ G(x)=e^{-\frac{1}{2}x^{T}\Sigma^{-1}x}$. For rendering, 3D Gaussians are projected onto 2D planes using a splatting method. This process involves a new covariance matrix $\Sigma^{\prime}$ in camera coordinates, which is given by: $\Sigma' = JW \Sigma W^TJ^T$, where $W$ is the viewing transformation matrix, and $J$ is the Jacobian of the affine approximation of the projective transformation. During training, the covariance matrix is optimized to minimize the difference between the rendered images and the ground truth.

\noindent\textbf{ControlNet.} ControlNet \cite{controlnet} is a deep learning architecture that enhances the controllability of generative models, aiming to precisely guide the generation process by adding additional control signals such as depth maps and normals. It incorporates a control network module into the generative model, allowing users to finely adjust specific features during image generation, such as expressions, poses, or other visual elements. In our work, we build a ControlNet conditioned on the motion information and landmarks of the source video to control the temporal consistency of the portrait editing results.

%% file: 4_method.tex
\section{Method}
\label{sec:method}
In this section, we present FYM, a generic optimization framework designed to ensure temporal consistency in portrait editing. An overview of the framework is provided in Fig. \ref{fig:pipeline}. Given a monocular video, we first apply 3D Gaussian splatting (3DGS) to render the source images. Next, we develop a diffusion model that effectively learns motion trajectories changes of pixel coordinates at multiple scales from the original video frames. We employ multi-resolution hash encoding to process the pixel coordinates and compute the differences between each frame's encoded result and the first frame. Additionally, to maintain fine-grained temporal consistency in facial expressions during portrait editing, we propose a dynamic re-weighted attention mechanism. This mechanism assigns higher weight coefficients to landmark coordinates and dynamically adjusts these weights based on landmark loss, resulting in more consistent and refined facial expressions.

\subsection{3DGS for Rendering Source Image}
\label{sec:3dgs}
Given a video $V$  comprised of frames $\{$${I_1, I_2, \ldots, I_K}$$\}$, which are aligned to a mesh template, specifically the deformed FLAME mesh \cite{flame}, we map 3D Gaussians onto the canonical mesh $\mathcal{M}$. Each Gaussian is characterized by parameters such as position $p$, rotation $r$, scale $s$, color $c$, and opacity $o$. These Gaussians act as semi-transparent 3D particles, rendered into camera views via point-cloud-based rasterization. Using SplattingAvatar \cite{splattingavatar}, we generate RGB renderings, which will serve as the images to be edited in the next stage, as shown below:
\begin{align}
\mathcal{F}_{SplattingAvatar}(I_i,\mathcal{M}(p,r,s,c,o))\to I_{i}^r \in R^{512 \times 512 \times 3}.
\end{align}

\subsection{Overall Content Temporal Consistency Module}
\label{sec:OCTCM}
In this subsection, we define the temporal consistency of the video as the consistency of the motion trajectories of pixel coordinates. For any foreground positions $p_{fg_i} = \{(x,y,i)\}$ in the $i^{th}$ frame of the rendered video $\{$${I_1^r, I_2^r, \ldots, I_K^r}$$\}$, we first encode it using a 3D multi-resolution hash encoding function $\mathcal{H}_{3D}(p_{fg_i})$, to obtain high-dimensional features. We then compute the difference between this feature and the corresponding positions in the first frame, yielding the motion trajectories of points $p_{fg_i} \in R^{512 \times 512}$. We assume the background is stationary, so the trajectory features of background positions are set to zero. Using these motion trajectories as a condition, we construct a diffusion model to reconstruct each frame of the rendered video. Next, when the user employs editing tools (e.g., IClights \cite{iclight}) to edit the frames $\{I_{i}^r\}$, the temporal consistency between frames may be disrupted, since the identity features may change after editing the portrait using editing tools. Therefore, we use the edited results to fine-tune the diffusion model from the previous stage, enabling it to adapt to the new character's features. In this way, the motion trajectories of the rendered video can guide the motion of the edited results.

\begin{figure}[h]
  \centering
    \includegraphics[width=0.8\linewidth]
    {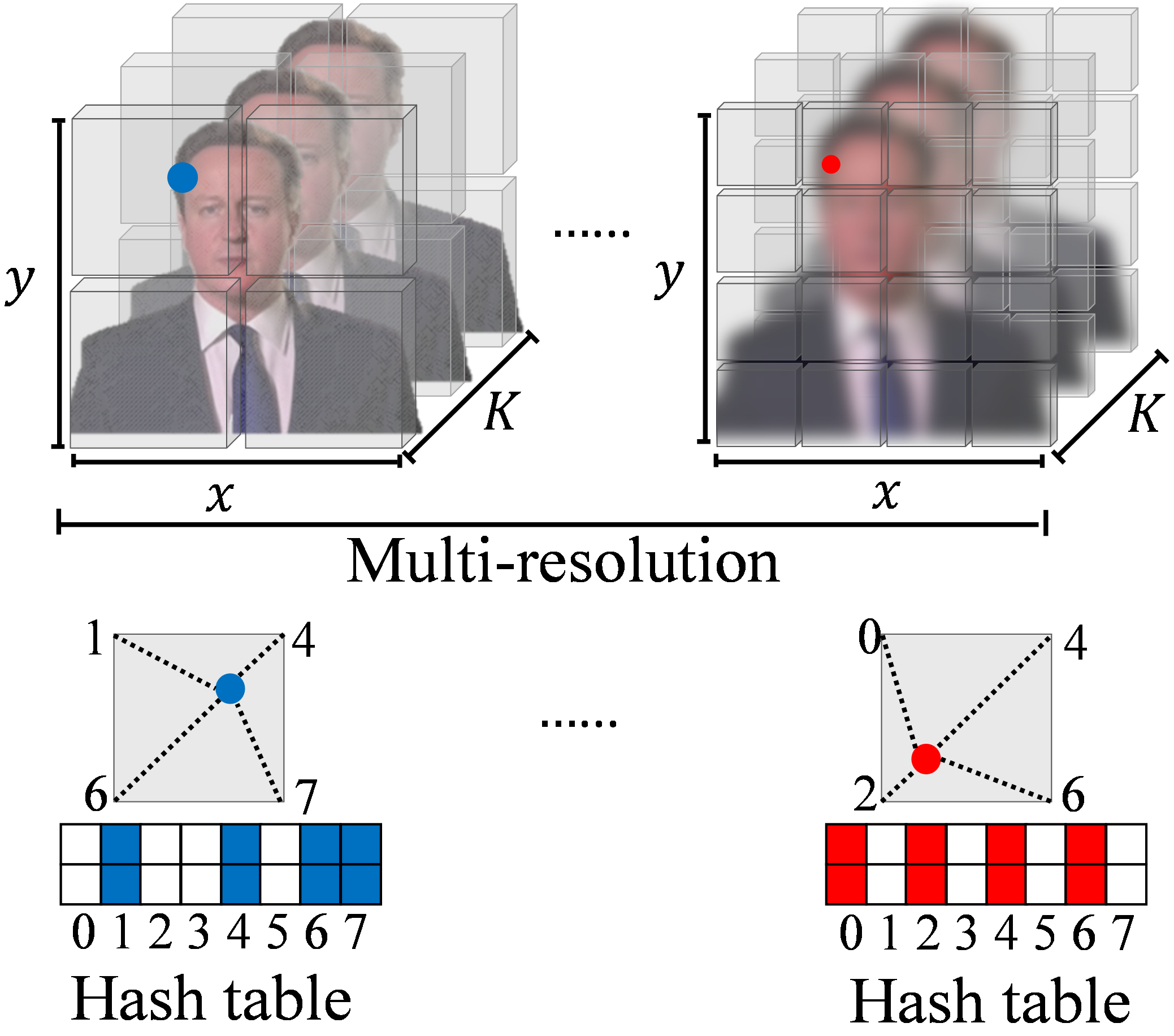}
    \caption{The schematic diagram of encoding spatial points using multi-resolution 3D hash encoding.}
    \label{fig:3Dhash}
\end{figure}

\noindent\textbf{Video Reconstruction Based on Diffusion Models and Motion Trajectories.}
As mentioned earlier, a point of $p_{fg_i}$ in the rendered video can be represented as a position in an orthogonal 3D space $(x,y,i)$. We use 3D hash encoding to represent the video space, which encapsulates it as an $L$-layer multi-resolution feature grid. The resolution of the $l^{th}$ layer is denoted as $R_l$. The feature grid refers to a grid populated with learnable features at each vertex, with a dimensionality of $F$:
\begin{align}
R_l=\lfloor R_{\min}\cdot n^l\rfloor,n=\exp\left(\frac{\ln R_{\max}-\ln R_{\min}}{L-1}\right),
\end{align}
when querying the $p_{fg_i}^l$ points at the $l^{th}$ layer, the input coordinates are scaled based on the grid resolution of that layer. The queried features of $p_{fg_i}^l$ are obtained through trilinear interpolation using the 8 neighboring corner points of the corresponding cube (see Fig. \ref{fig:3Dhash}), with the corner points determined by applying rounding down and up operations to the $p_{fg_i}^l$ coordinates:
\begin{align}
\lfloor p_{fg_i}^l\rfloor=\lfloor p_{fg_i}\cdot R_l\rfloor,\lceil p_{fg_i}^l\rceil=\lceil p_{fg_i}\cdot R_l\rceil,
\end{align}
we construct a feature vector array of size $T$ to store the features mapped from each corner point to the corresponding features of the layer. When querying the hash-encoded feature of $p_{fg_i}$, the mapping is performed as follows:
\begin{align}
\mathcal{H}_{3D}(p_{fg_i})=
\begin{pmatrix}
\oplus_{i=1}^dx_i\pi_i
\end{pmatrix}{\mathrm{mod}}\,\,T,
\end{align}
where $\oplus$ represents the bit-wise XOR operation and $\pi_i$ are unique large prime numbers following \cite{ing}. Then we calculate the coordinate encoding difference between the current frame and the first frame to represent the trajectories:
\begin{align}
\mathcal{TRA}_i=\mathcal{H}_{3D}(p_{fg_i})-\mathcal{H}_{3D}(p_{fg_1}) \space,i \in [2, K].
\end{align}

\begin{figure*}[h]
\begin{align}
    \begin{aligned}
\mathcal{L}_{CFM} & 
=\mathbb{E}_{{t,p_{t}(z|\epsilon),p(\epsilon),\mathcal{TRA}_i,I_{1}^r}}\left(-\frac{b_t}{2}\lambda_t^{\prime}\right)^2||\epsilon_\Theta(z,t,\mathcal{TRA}_i,I_{1}^r)-\epsilon||_2^2,\epsilon_\Theta:=\frac{-2}{\lambda_t^{\prime}b_t}(v_\Theta-\frac{a_t^{\prime}}{a_t}z).
\end{aligned}
\end{align}
\end{figure*}
Next, we build a trajectory-conditioned diffusion model to render the reconstruction of each frame in the video with SD3 \cite{SD3}. Unlike traditional diffusion models, SD3 does not simulate the random process from data to noise. Instead, it introduces a new generative model formulation—Histogram Flow. To express the relationship between noisy data $z_t$, the original data $x_0$, and noise $\epsilon$, SD3 defines two functions: $\psi_{t}(\cdot|\epsilon)$ and $u_{t}(z|\epsilon)$. Specifically, $\psi_{t}$ describes the mapping from $x_0$ to $a_{t}x_{0}+b_{t}\epsilon$, while $u_t$ represents the derivative of this mapping:
\begin{equation}
\begin{aligned}
    \psi_{t}(\cdot|\epsilon):x_{0}\mapsto a_{t}x_{0}+b_{t}\epsilon, \\
    u_{t}(z|\epsilon):=\psi_{t}^{\prime}(\psi_{t}^{-1}(z|\epsilon)|\epsilon).
\end{aligned}
\end{equation}
$u_{t}(z)$ is a marginal vector field, which can be constructed through the conditional vector field $u_{t}(z|\epsilon)$. These conditional vector fields generate the marginal probability path $p_{t}(z)$. This vector field is defined through the expectation, considering all possible noise $\epsilon$. The loss function for the entire training process minimizes the difference between the velocity field $v_{\Theta}(z,t)$ predicted by the neural network and the actual velocity field $u_{t}(z)$, and introduces conditional flow matching to provide a feasible objective $\mathcal{L}_{CFM}$ (see eq.(6)), where $t=0, \cdots, T-1$ represents the time step in the diffusion process. $I_1^r$ is the first frame of the rendered portrait video.

\noindent\textbf{Fine-tuning the trajectory-conditioned diffusion model using edited portraits.} Our work enhances the output of several advanced portrait editing tools, such as IClights \cite{iclight} and InstructPixel2Pixel \cite{instructpix2pixl}. By processing rendered images with these tools, we can obtain results $\{$${I_1^e, I_2^e, \ldots, I_K^e}$$\}$. However, directly inputting these edited portraits into the model from the previous stage is not effective, as the identity features of the edited portraits differ from those in the original rendered images. To address this, We propose a fine-tuning-based ControlNet adaptation method, which fine-tunes ControlNet on the target portrait dataset to enable it to adapt to the new portrait's features. We employ the same method as described in Sec \ref{sec:OCTCM} to achieve the reconstruction of edited portraits. Using 3D hash multi-resolution encoding, we obtain the pixel coordinates of the edited portraits and the motion trajectories from the first frame of the edited portraits, which are then used as conditions to fine-tune ControlNet. The loss function can be expressed as follows, where $I_1^e$ is the first frame of the edited portrait video.
\begin{align}
\mathcal{L}_{CFM} & 
=\mathbb{E}\left(-\frac{b_t}{2}\lambda_t^{\prime}\right)^2||\epsilon_\Theta(z,t,\mathcal{TRA}_i^e,I_{1}^e)-\epsilon||_2^2,
\end{align}

 \subsection{Dynamic Re-weighted Attention Mechanism}
 \label{sec:dram}
 After the Overall Content Temporal Consistency Module, although the edited portrait generally follows the motion trajectories of the rendered video, the fine-grained facial expression control is still not ideal. Therefore, we propose an attention dynamic re-weighting mechanism, which updates the weight factors of the landmark coordinates using a landmark loss to achieve precise control over facial expressions.
 \noindent\textbf{Weight factor initialization.} Before training the network, we construct a weight matrix $\mathcal{W}_{512\times512} = \textbf{1}$, which means that during the cross-attention process, the influence of all coordinate motion trajectories is equal, and the weight factors are all set to $1$.
 
 \noindent\textbf{Portrait landmarks detection.} We employ EMOCA\cite{emoca} to detect the landmarks of both the rendered portrait $I_i^r$ and the predicted edited portrait $I_i^{e'}$, as shown below:
 \begin{align}
     \textbf{ld}_i = \mathcal{EMOCA}(I_i^{r/e'}),
 \end{align}
 where  $\textbf{ld}_i$ is the set of face landmarks detected by EMOCA. The coordinates of each landmark point $(x_m,y_n)$ represent the corresponding 2D spatial location in the $i^{th}$ frame of the image.

\noindent\textbf{Dynamic weight factor update.} During the network's iterative training process, when the diffusion model denoises to obtain the predicted edited portrait, we use the landmark loss to calculate the difference between it and the rendered portrait for each landmark point, we use Euclidean distance to calculate the differences for each landmark point, as follows:
\begin{align}
\mathcal{L}_{ld_{imn}} = (ld_i^r(x)_m-ld_i^{e'}(x)_m)^2 + (ld_i^r(y)_n-ld_i^{e'}(y)_n)^2
\end{align}
When the loss of a point is greater, it means that the point requires a higher attention weight factor. We directly take this loss as part of the weight factor. Therefore, the motion trajectory and weight factor are updated as follows:
\begin{align}
   \mathcal{TRA}_{imn} =  \mathcal{TRA}_{imn} \times (1 + \mathcal{L}_{ld_{imn}}).
\end{align}

%% file: 5_experiment.tex
\begin{figure*}[h]
  \centering
    \includegraphics[width=0.95\linewidth]
    {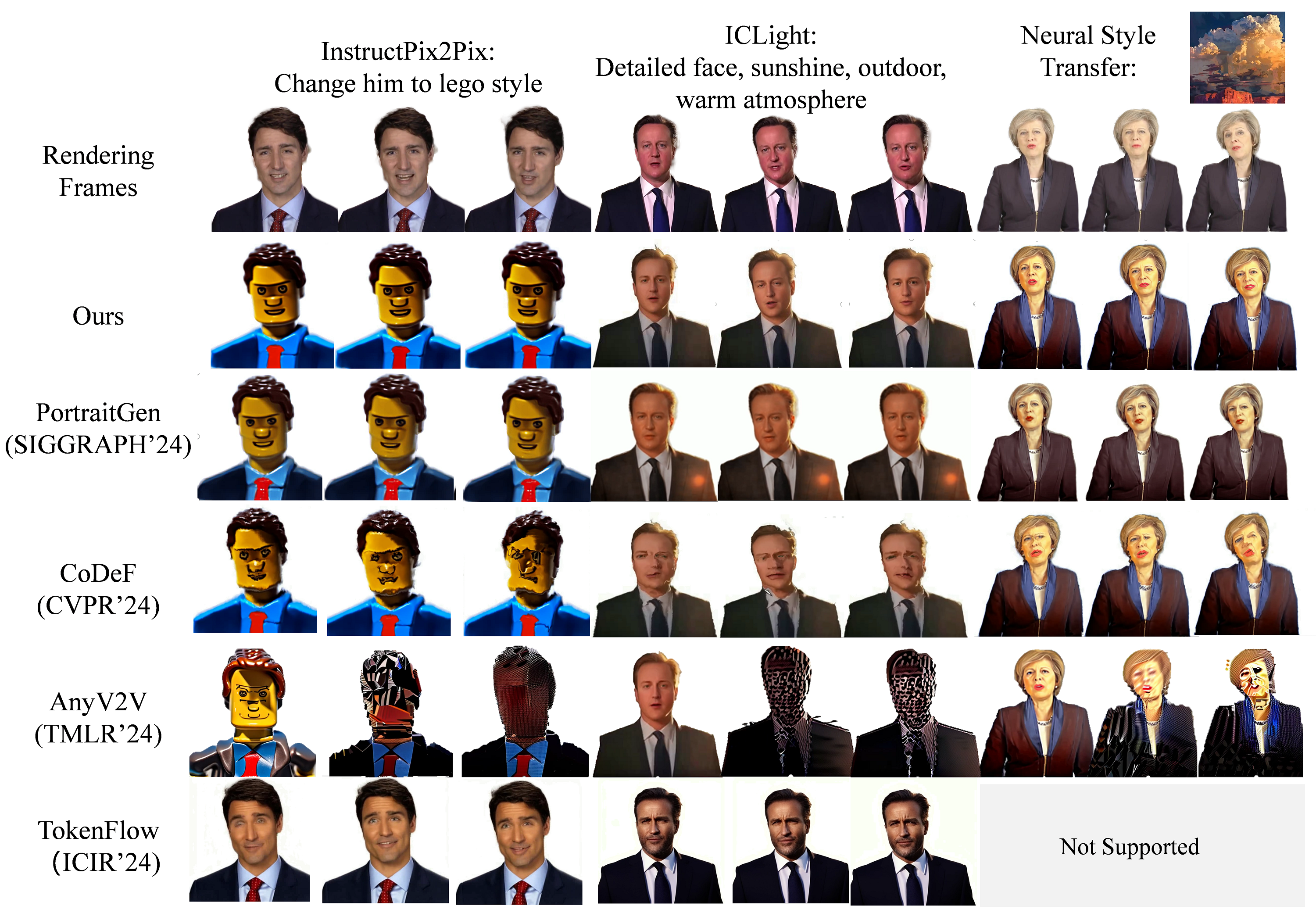}
    \caption{Quantitative results. Compared to state-of-the-art methods, our approach is able to generate high-quality, temporally consistent editing results that align with the prompt. For a clearer comparison of the temporal consistency across different methods\textbf{(you can enlarge the image to make a comparison)}. You can refer to the demo in the supplementary materials.}
    \label{fig:results}
\end{figure*}

\section{Experiment}
\label{sec:exp}
\subsection{Setup}
\textbf{Datasets.} We employ the dataset released by GAO et al. \cite{portrait}, which includes a total of 8 characters. These datasets, collected from the internet, consist of monocular videos of the upper body, with each character featuring approximately one minute of speaking video along with corresponding tracking parameters. We carefully use the last 480 frames for Gaussian splatting validation, subsequent editing operations, and training of the diffusion model. The remaining frames are used for training 3D Gaussian splatting.

\noindent\textbf{Evaluation Criteria.} We use CLIP Text-Image Direction Similarity (${CLIP}{dirs}$) and CLIP Direction Consistency (${CLIP}{dirc}$) to assess the alignment of 3D edits with text instructions. Additionally, we conduct a user study to evaluate video quality across four dimensions: Text Alignment (TA), Fidelity (FL), Temporal Consistency (TC), and Overall Impression. Participants first view the original video to understand the editing goals, then select the best video for each criterion in each case. 

\noindent\textbf{Implementation Details.} Our framework is implemented in PyTorch and runs on an RTX 3090 GPU. We use Stable Diffusion version 3, which consists of two stages. In the first stage, we strictly follow the training procedures of SplattingAvatar\cite{splattingavatar}. In the second stage, we train a diffusion model conditioned on motion trajectories. The Adam optimizer is used with a learning rate of $10^{-4}$, performing 5,000 iterations with a batch size of 4. The feature dimension $F$ of the multi-resolution hash encoding is set to 2, and the number of resolution layers $L$ is set to 16.

\subsection{Qualitative Evaluation}
We compare our method with state-of-the-art video editing techniques, including PortraitGen \cite{portrait}, CoDeF \cite{codef}, TokenFlow \cite{tokenflow}, and AnyV2V \cite{anyv2v}. Qualitative comparisons of text-driven and image-driven editing are presented in Figure 3. For PortraitGen, we generate higher-quality, high-resolution edited results, whereas the quality of images produced by PortraitGen is limited due to the Gaussian neural rendering it employs. CoDeF performs well in maintaining temporal and detail consistency in short, slow-motion videos, but struggles when facial expressions and postures change rapidly. TokenFlow, on the other hand, often fails to align edits with the given prompts due to the extended attention mechanism, which causes the latent codes to drift out of the domain and degrade the results. AnyV2V shows instability in portrait editing, likely because its temporal layers do not provide sufficient contextual constraints. In contrast, our method consistently generates high-quality, prompt-aligned, and temporally consistent edits. For a clearer comparison of temporal consistency, please refer to the video in the supplementary materials.

\subsection{Quantitative Evaluation}
We calculate the $CLIPdirs$ and $CLIPdirc$ for each method on $10$ edited videos ($5$ characters, each with $2$ edited videos). To evaluate the temporal consistency of the editing results, we calculate the optical flow variation between frames of the edited result and compare it with the optical flow variation between frames of the source rendering result. Experiments show that the closer the optical flow variation between frames of the edited result is to that of the source rendering result, the better the temporal consistency. We perform optical flow calculations on the 10 edited videos mentioned above and take the average. The results are shown in Tab. \ref{tab:consistency1}
. It can be observed that both our method and PortraitGen exhibit good temporal consistency, but our method achieves \textbf{higher quality} compared to PortraitGen. Additionally, we invite 20 graduate and doctoral students from related fields to participate in a user study, evaluating four metrics: text alignment, fidelity to the source video, temporal consistency, and overall impressions. The results are shown in Tab. \ref{tab:consistency2}. For each group of editing results, participants addressed the following queries:

\noindent Q1. Which video best matches the text description?
 
\noindent Q2. Which video is the most realistic?
 
\noindent Q3. Which video has the best temporal consistency?
 
\noindent Q4. Which video is the best overall?

\begin{table*}[t!]
    \centering
    \setlength{\tabcolsep}{0.5mm}
        
        \vskip 0.15in
    \resizebox{0.65\linewidth}{!}{
    \begin{tabular}{c|c|c|c|c|c|c|c|c|c|c|c}
        Method\ /\ Frame & 1-2 & 2-3 & 3-4 & 4-5 & 5-6 & 6-7 & 7-8 & 8-9 & 9-10 & 10-11 & total error\\
        \midrule
         Rendering results & 0.22 & 0.46 & 0.23 & 0.03 & 0.23 & 0.17 & 0.12 & 0.20 & 0.07&0.04 & 0\\
         PortraitGen& 0.13 & 0.27 & 0.05 & 0.19 & 0.12 & 0.04 & 0.05 & 0.26 & 0.04 &0.10 & \cellcolor{green} 0.52 \\
         CoDeF & 0.08 & 0.14 & 0.12 & 0.10 & 0.11 &0.09 & 0.08 & 0.09 & 0.07 & 0.08 & 0.81\\
         AnyV2V & 0.78 & 0.08 & 0.09 & 0.31 & 0.13 & 1.36 & 1.17 & 2.80 & 2.03& 2.81& 9.79\\
         TokenFlow & 0.79 & 0.25 & 0.28 & 0.27 & 0.19 & 0.06 & 0.47 & 0.32 & 0.21& 0.14& 1.21\\
        \textbf{Ours} & 0.21 & 0.25 & 0.15 & 0.06 & 0.13 &0.09 & 0.08 & 0.16 & 0.08 & 0.03& \cellcolor{green}0.53\\
 
        \bottomrule
        
    \end{tabular}}
    \caption{
        Quantitative comparisons on the temporal consistency of editing outcomes with other methods. The closer the magnitude of optical flow variation between frames is to the optical flow variation magnitude of the rendering results, the better the temporal consistency.
        }
    \label{tab:consistency1}
    \vskip -0.1in
\end{table*}

\begin{table}[t!]\small
    \centering
   \setlength{\tabcolsep}{0.5mm}
    \begin{tabular}{c|c|c|c|c|c|c}
        \multirow{2}{*}{Method} & \multicolumn{2}{c|}{Objective Metrics}   &\multicolumn{4}{c}{User Study} \\ 
        &CLIPdirs$\uparrow$&CLIPdirc$\uparrow$&TA$\uparrow$&FL$\uparrow$&TC$\uparrow$&Overall$\uparrow$\\
        \midrule
        PortraitGen & 0.2924 &99.31 & 40\% & 40\% & 20\% & 5\% \\
        CoDeF & 0.2792 & 97.88 & 0\% & 0\% & 0\% & 0\%  \\
        AnyV2V & 0.2736 & 97.21 & 0\% & 0\% & 0\% & 0\% \\
        TokenFlow & 0.2897 & 98.63 & 0\% & 10\% & 5\% & 0\% \\
        \textbf{Ours} & \cellcolor{green}\textbf{0.3007} & \cellcolor{green}\textbf{99.53} & \cellcolor{green}\textbf{60\%} & \cellcolor{green}\textbf{50\%} & \cellcolor{green}\textbf{75\%} & \cellcolor{green}\textbf{95\%} \\
        \bottomrule
    \end{tabular}
    \caption{
        Quantitative comparisons on the quality of editing outcomes with state-of-the-art methods.
        }
    \label{tab:consistency2}
\end{table}

\subsection{Ablation Study}
To evaluate the various components of our method, we examine the following aspects: 1) The influence of the multi-resolution hash encoding on the final results. 2) The impact of dynamic re-weighted attention mechanism on the final editing outcomes.

\begin{figure}[h]
  \centering
    \includegraphics[width=0.95\linewidth]
    {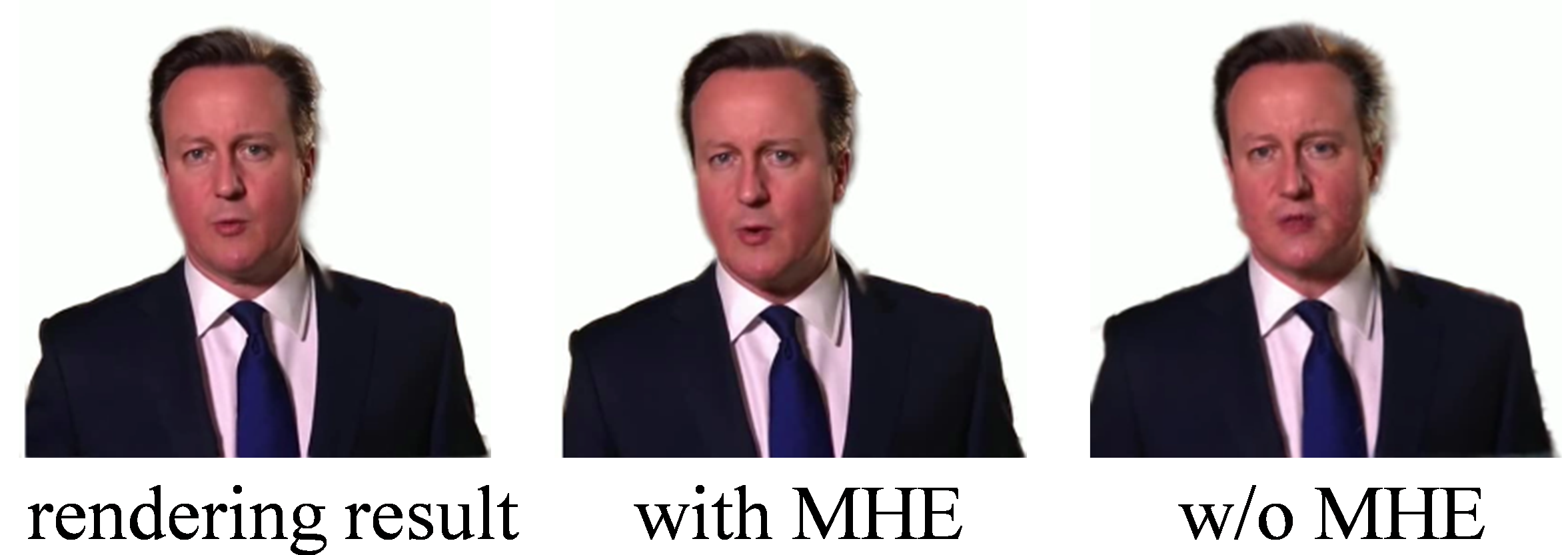}
    \caption{Ablation experiment results of Multi-resolution Hash Encoding (MHE).}
    \label{fig:mhe}
\end{figure}

\noindent\textbf{Multi-resolution Hash Encoding (MHE).} We propose using multi-resolution hash encoding to better learn the facial details and motion trajectories of pixel coordinates during the video reconstruction stage. We set the number of resolution layers to $1$, which results in motion estimation learning at a single scale, similar to flow-based methods. We compare this case with our method, and the experimental results are shown in Fig. \ref{fig:mhe}. It can be observed that when MHE is used, the reconstructed result is closer to rendering result.

\begin{figure}[t]
  \centering
    \includegraphics[width=\linewidth]
    {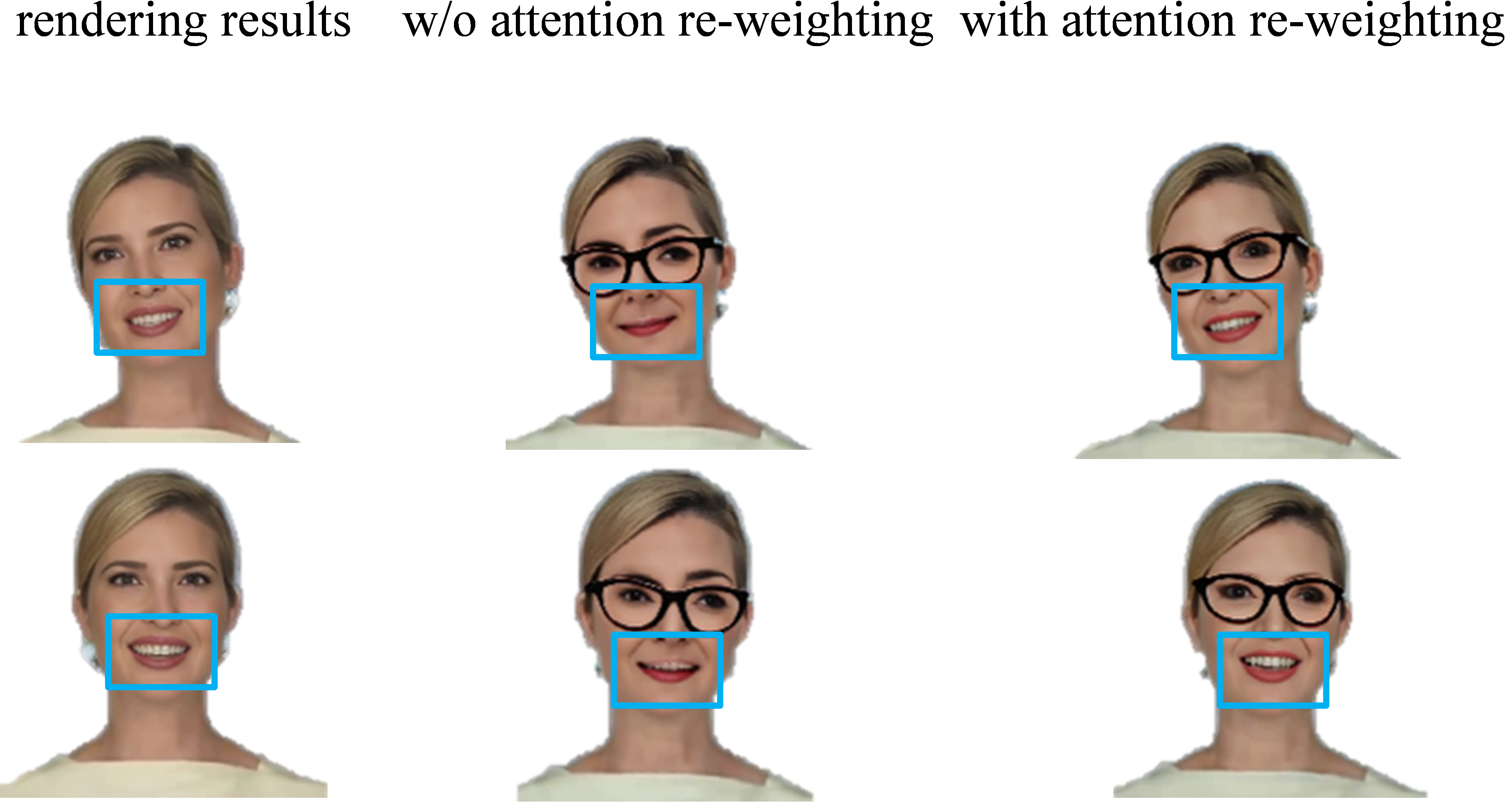}
    \caption{Ablation qualitative evaluation results of Dynamic Re-weighted Attention Mechanism (DRAM). Prompt: Give her a pair of glasses.}
    \label{fig:dram}
\end{figure}

\noindent\textbf{Dynamic Re-weighted Attention Mechanism (DRAM).} To better adapt to the portrait editing task, we propose an attention dynamic re-weighting mechanism to compensate for the issue of facial expression inconsistency during the editing process. We conduct specific experiments comparing DRAM with and without its application to verify whether the attention mechanism truly constrains the portrait expressions. As shown in Fig. \ref{fig:dram}, it can be observed that after adding DRAM, the generated portrait expressions are closer to the rendering result (pseudo-ground truth, obtained from Gaussian rendering in the first stage). Additionally, we also introduce an expression error (EE), using EMOCA \cite{emoca} to calculate the error between the expressions in the two cases and the pseudo-ground truth, as shown in Tab. \ref{tab:ee}.
\begin{table}[t!]
    \centering
    \setlength{\tabcolsep}{1mm}  
    \resizebox{0.3\linewidth}{!}{
    \begin{tabular}{c|c}
        Method & EE$\downarrow$ \\
        \midrule
         w/o DRAM & 2.27\\
        \textbf{with DRAM} & \cellcolor{green} \textbf{1.15} \\
        \bottomrule
    \end{tabular}}
         \caption{
        Ablation study experiment results of Dynamic Re-weighted Attention Mechanism (DRAM).}
        \label{tab:ee}
\end{table}

%% file: 6_discussion.tex
\section{Discussion}
\label{sec:discussion}
\begin{figure}[t]
  \centering
    \includegraphics[width=\linewidth]
    {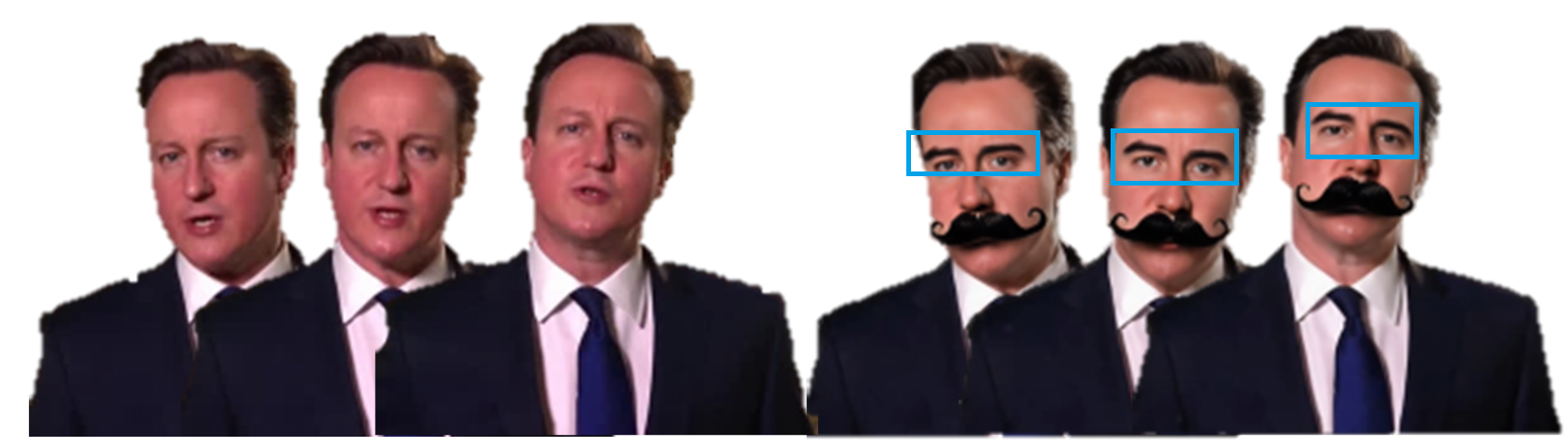}
    \caption{The experimental results of edit discomfort. Prompt: Give him a mustache.}
    \label{fig:limi}
\end{figure}

Our method provides a novel yet intuitive approach to optimizing the temporal consistency of portrait editing. However, there are many other excellent methods, such as the one-shot method AnyV2V \cite{anyv2v}, which better cater to user preferences, although they may not perform as well in terms of quality. Our method has shown good results with portraits, so applying it to general scenes seems feasible. However, the temporal consistency in scenes with more detailed variations will pose a challenge, which will be a direction for our future research. Our method also has the following limitations: 1) As mentioned above, our method requires training for each edit, making it edit-specific. 2) The final presentation depends on the robustness of the editing tools; it can only solve temporal consistency issues but cannot optimize issues with the edits themselves, as shown in Fig \ref{fig:limi}.

%% file: 7_conclusion.tex
\section{Conclusion}
\label{sec:conclusion}
We propose a framework called "Follow Your Motion" (FYM) to ensure temporal consistency in portrait editing. Given portrait images rendered by a pre-trained 3D Gaussian Splatting model, we first design a diffusion model that intuitively learns motion trajectory changes at different scales and pixel locations, covering the transition from the first frame to each subsequent frame. Furthermore, to maintain fine-grained expression temporal consistency in portrait editing, we introduce a dynamic re-weighted attention mechanism. This mechanism assigns higher weights to landmarks in space and dynamically adjusts these weights based on landmark loss, achieving more consistent and refined facial expression representation.